\documentstyle{article}

\newcommand{\lgg}{{\rm lgg}}

\renewcommand{\aa}{\underline}
\newcommand{\oo}{\overline}

\renewcommand{\phi}{\varphi}
\renewcommand{\epsilon}{\varepsilon}

\newcommand{\calA}{{\cal A}}

\newcommand{\calE}{{\cal E}}

\newcommand{\calG}{{\cal G}} 
\newcommand{\calH}{{\cal H}}

\newcommand{\calM}{{\cal M}}

\newcommand{\calP}{{\cal P}}  

\begin{document}

\begin{center}
{\Large\bf The Prioritized Inductive Logic Programs}\footnote[1]{{The
project was partially supported by the National Natural Science Foundation of
China and the National 973 Project of China under the grant number
G1999032701.}}\\[0.5cm]
{Shilong Ma$^{\dag,}$\footnote[2]{{Email: \{slma,
kexu\}@nlsde.buaa.edu.cn.}}, Yuefei Sui$^{\ddag,}$\footnote[3]{{Email: suiyyff@hotmail.com.}}, Ke Xu$^{\dag ,2}$}\\[0.5cm]
{\small {}$^{\dag }$ Department of Computer Science \\[0pt]
Beijing University of Aeronautics and Astronautics, Beijing 100083, China}\\[0pt]
{\small {}$^{\ddag }$ Key Laboratory of Intelligent Information Processing\\
Institute of Computing Technology, Chinese Academy of Sciences, Beijing 100080, China}
\bigskip

\begin{minipage}{11cm} 
{\footnotesize {\bf   Abstract}: The limit behavior of inductive logic programs has not been explored, but when 
considering incremental or online inductive learning algorithms which usually run ongoingly, such behavior of the programs 
should be taken into account. An example is given to show that some inductive learning algorithm may not be correct in the long 
run if the limit behavior is not considered.
An inductive logic program is {\it convergent} if given an increasing sequence of example sets, the program 
produces a corresponding sequence of the Horn logic programs which has the set-theoretic limit,
 and is {\it limit-correct} if the limit of the produced sequence of the Horn logic programs is correct
with respect to the limit of the sequence of the example sets. It is shown that the GOLEM system is not
limit-correct. Finally, a limit-correct inductive logic system, called  the prioritized GOLEM system, is proposed as
a solution.

}
\end{minipage}
\end{center}

\bigskip
\bigskip

\noindent{\bf 1. Introduction}
\smallskip

 As information increases exponentially, it becomes more and more important to 
 discover useful knowledge in massive information. Inductive logic programming
 is used in learning a general theory from given examples. In incremental
 learning, the examples are usually given one by one. After a new example is obtained, 
 the current theory learned 
 from previous examples might need to be updated to fit all the examples given so far. 
 Thus we get a sequence of theories $\Pi _{1},\Pi
_{2},\cdots ,\Pi _{n},\cdots $. Sometimes there might be infinitely many examples 
and this procedure may not stop, i.e., there may not exist a natural number $k$ such that $\Pi _{k}=\Pi
_{k+1}=\cdots $. For example, if we restrict the theories to Horn logic
programs, then there exists some Herbrand interpretation $I$ such that we
will never find a finite program $\Pi $ whose least Herbrand model is equal to $I,$
because the set of Herbrand interpretations is uncountable while the
set of finite Horn logic programs is only countable ([10]). So we should
consider some kind of the limits of theories which 
should be correct with respect to all the examples.

Formally introducing the set-theoretic limits of sequences of first order theories into logic
and computer science, and using theory versions as approximations of some
formal theories in convergent infinite computations are the independent
contributions by Li. Li ([4],[5]) defined the set-theoretic limits of
first order theories, and thereon gave a formal system of the inductive
logic. Precisely, given a sequence $\{\Pi _{n}\}$ of the first order
theories, the set-theoretic limit of $\{\Pi_n\},$ denoted by $\Pi =\displaystyle\lim_{n\rightarrow
\infty }\Pi _{n},$ is the set of the sentences such that every sentence in $%
\Pi $ is in almost every $\Pi _{n},$ and every sentence in infinitely
many $\Pi _{n}$'s is in $\Pi $ also. The set-theoretic limit does not always exist
for any sequence of the first order theories. In the following sections we use the limits as the set-theoretic limits.

The limit behaviors of an inductive logic program should be an important research topic which has not been explored.
Nowadays most of the softwares and algorithms are incremental or online, which run in the long run. 
When we consider the correctness problem of such softwares and algorithms, their limit behaviors should be taken into 
account. 
In this paper we focus on incremental inductive learning algorithms in inductive logic programs. Assume that examples come in sequences,
let $E_n$ be the example set at time $n.$ An inductive learning algorithm $\calA$ produces a theory $\calA(E_n),$ for every $n,$ which is correct with 
respect to $E_n.$ Later we shall give an example to show that some inductive learning algorithms may not be correct in the long 
run if the limit behavior is not considered. Concerning the limit behavior, a reasonable inductive learning algorithm 
$\calA$ should satisfy the following conditions:

$\bullet$ {\it The convergence}: Given a sequence $\{E_n\}$ of the example sets such that $E_1\subseteq E_2\subseteq E_3\subseteq \cdots,\ 
\{\calA(E_n)\}$ has the set-theoretic limit;

$\bullet$ {\it The limit-correctness}:  The limit of
$\{\calA(E_n)\}$ should be correct with respect to the limit of $\{E_n\},$ that is, for any $e\in \lim_{n\rightarrow\infty}
E_n,\ \lim_{n\rightarrow\infty} \calA(E_n)\vdash e;$

Based on the above requirements, we consider the empirical ILP systems: FOIL, GOLEM and MOBAL. Because FOIL and MOBAL are function-free,
we focus on the GOLEM system.

In the following, our discussion is based on a fixed logical language which contains only finitely many predicate symbols.

The authors ([7]) considered the limit behavior of the Horn logic programs, and proved the following theorem:

{\bf Theorem 1.1}([7]). Given a sequence $\{\Pi _{n}\}$ of Horn logic programs, if $\Pi =\displaystyle%
\lim_{n\rightarrow \infty }\Pi _{n}$ exists and for every sufficiently large $n,\ \Pi_n$  satisfies an assumption that 
 {\it for every clause in $\Pi_n$, every term occurring in the body also occurs in the head,}
  then 
  $$\lim_{n\rightarrow\infty}\calM(\Pi_n)=\calM(\lim_{n\rightarrow\infty}\Pi_n),$$ 
  where $\calM$ is an operator such that
  for any Horn logic program $\Pi,\ \calM(\Pi)$ is the least Herbrand
  model of $\Pi.$
  
To consider the limit-correctness of inductive logic programs, we assume that the inductive logic programs satisfy
the convergence. Given an inductive logic program $\calA$, if for every positive example set $E_n,\ \calA(E_n)$ is a Horn logic program 
which is correct with respect to $E_n,$ i.e., $\calM(\calA(E_n))\supseteq E_n,$ and $\calA(E_n)$ satisfies the assumption, 
then by theorem 1.1, $$\calM(\lim_{n\rightarrow\infty}\calA(E_n))=
\lim_{n\rightarrow\infty}
\calM(\calA(E_n))\supseteq \lim_{n\rightarrow\infty} E_n,$$ 
so  
$\calA$ is limit-correct. Therefore, to make $\calA$ satisfy the limit-correctness, we should design $\calA$ such that for any input $E,
\ \calA(E)$ is a Horn logic program satisfying the assumption. We shall give two examples to show that the current 
GOLEM system may not always produce a logic program satisfying the assumption. We modify the GOLEM system to be a prioritized one in which
a priority order is defined on the literals. In detail, let
$\calG$ and $\calP$ be the GOLEM algorithm and the prioritized GOLEM algorithm such that for any example set $E,\ \calG(E)$ 
and $\calP(E)$ are the Horn logic program produced by the GOLEM system and the prioritized GOLEM system, respectively.
Then, for a set $E$ of examples,  $\calG(E)$ may not satisfy the assumption, but $\calP(E)$ satisfies the assumption. Hence, $\calP$ 
is limit-correct.

The paper is organized as follows. In section 2, we shall give the basic definition of the GOLEM system,  the distance
on terms and formulas, and the set-theoretic limits.
In section 3, we shall consider the limit-correctness of the GOLEM system, 
and give two examples to show that the GOLEM system is not limit-correct and sensitive to the ordering of the examples.
In section 4, we shall propose the prioritized GOLEM system, and prove that the prioritized GOLEM system is limit-correct,
and not sensitive to the ordering of the examples. The last section concludes the paper.

Our notation is standard, references are [1,2,6].

\bigskip
\noindent{\bf 2. The GOLEM system}

\smallskip 

In this section, we first give a basic introduction to the GOLEM system.
Similar to Nienhuys-Cheng's definition of the distance on terms and formulas([10]), we give the following definition.

{\bf Definition 2.1.} Let $f$ and $g$ be an $n$-ary and
an $m$-ary function symbols, respectively. The distance $\rho$ is defined as follows.

(2.1)\ $\rho(t, t)=0$, for any term $t;$

(2.2)\ If $f \not= g$, then $\rho(f(t_1,...,t_n), g(s_1,..., s_m))=1;$

(2.3)\ $\rho(f(t_1,..., t_n),
f(s_1,...,s_n))=\displaystyle\frac{\max\{\rho(t_i,
s_i)\mid 1\leq i\leq n\}}{\max\{\rho(t_i, s_i)\mid 1\leq
i\leq n\}+1},$

\noindent where $t_1,...,t_n, s_1,...,s_m$ are terms.

The distance defined above is a little
different from the one given by Nienhuys-Cheng in that the value which the
distance can take has a simple form $\displaystyle\frac{1}{m}$ for some
natural number. Such a distance is used to measure the difference between two
trees in graph theory. Every term $t$ can be taken as a tree $T_t.$ For
example, $t=f(t_1,...,t_n),$ the tree $T_t$ has a root with symbol $f$ and $%
n $-many children $T_{t_1},...,T_{t_n}.$ We say that two terms $t$ and $%
t^{\prime}$ are the same to depth $m$ if $T_t$ and $T_{t^{\prime}}$ are the
same to depth $m$. 
\bigskip

Given two clauses $C_1,C_2,$ to compute the least general generalization of $C_1$ and $C_2,$ denoted by $\lgg(C_1,C_2),$ 
we give the following procedure:

Step 1. Given two terms $t=f(t_1,...,t_n)$ and $s=g(s_1,...,s_m),$
$$\lgg(t,s)=\left\{ \begin{array}{ll}
v &\mbox{if $f\ne g$}\\
f(\lgg(t_1,s_1),...,\lgg(t_n,s_n)) &\mbox{if $f=g,$}
\end{array} \right. $$
where $v$ is any new variable.

Step 2. Given two literals $l_1=(\lnot)^{k_1} p(t_1,...,t_n)$ and $l_2=(\lnot)^{k_2} q(s_1,...,s_m),$
$$\lgg(l_1,l_2)=\left\{ \begin{array}{ll}
\mbox{undefined} &\mbox{if $k_1\ne k_2$ or $p\ne q$}\\
(\lnot)^{k_1}p_1(\lgg(t_1,s_1),...,\lgg(t_n,s_n)) &\mbox{if $k_1=k_2$ and $p=q,$}
\end{array} \right. $$
where $k_1=0$ or 1, $(\lnot)^0=$ \empty, $(\lnot)^1=\lnot.$

Step 3. Given two clauses $C_1=\{l_{1},...,l_{n}\}$ and $C_2=\{l'_{1},...,l'_{m}\},$
$$\lgg(C_1,C_2)=\{\lgg(l_{i},l'_{j}): 1\leq i\leq n, 1\leq j\leq m, \lgg(l_{i},l'_{j})\mbox{\ is\ defined}\}.$$

\bigskip
Given two sets $A, B$ of clauses, define 
$$\begin{array}{rl}
\lgg(A,B)=\{\lgg(C_1,C_2): & C_1 \in A,C_2\in B, d(C_1,C_2)=\min\{d(C_1,C_2): C_2\in B\}\\
& \&\ \lgg(C_1,C_2)\ \mbox{exists}\}.
\end{array}$$ 
Muggleton and Feng [8] showed that if $\Gamma$ is a finite set of ground literals then the rlgg of two clauses of $C_1$ and $C_2$ with respect to $\Pi$ is the lgg of $\Pi\rightarrow C_1$ and $\Pi\rightarrow C_2.$ Based on the property, they gave the ILP learning system GOLEM which is the only learning system explicitly based on the notion of relative least general generalization.

{\bf The GOLEM system:}

Suppose we are given a logic program $\Gamma$ (i.e., a background knowledge) and two examples (two ground atoms) 
$E_1$ and $E_2$ such that $\Gamma\not\vdash E_1$ and $\Gamma\not\vdash E_2.$ We construct the lgg $C$ of $E_1$ 
and $E_2$ relative to $\Gamma,$ so $\Gamma\land C\vdash E_1\land E_2$ and $C$ is used only once in the derivation 
of both $E_1$ and $E_2.$ Let $\Gamma=\{p_1,...,p_n,...\}.$ 

Define $C_1=((\lnot p_1\lor \lnot p_2\lor \cdots)\lor E_1),  C_2=((\lnot p_1\lor \lnot p_2\lor \cdots)\lor E_2),$ 
set $C=\lgg(C_1,C_2).$

\bigskip

{\bf Definition 2.2.} Given a sequence $\{A_n\}$ of sets of formulas, the set-theoretic limit of 
$\{A_n\}$ exists, denoted by $\lim_{n\rightarrow\infty} A_n,$ if 
$$\oo{\lim}_{n\rightarrow\infty} A_n=\aa{\lim}_{n\rightarrow\infty} A_n$$ 
where 
$$\begin{array}{l}
\oo{\lim}_{n\rightarrow\infty} A_n=\{\phi: \exists^\infty n(\phi\in A_n)\};\\
\aa{\lim}_{n\rightarrow\infty} A_n=\{\phi: \exists n_0\forall n\ge n_0(\phi\in A_n)\},
\end{array}$$
where $\exists^\infty n$ means that there are infinitely many $n.$

\bigskip
\noindent{\bf 3. The limit-correctness of the GOLEM system}

\smallskip

In this section we consider the limit-correctness of the GOLEM system, and give two examples to show that 
the GOLEM system is not limit-correct, and sensitive to the ordering of the examples. 
\bigskip

 Let $p$ be a predicate saying that $x$ is an even number if $p(x);$ and $s$ be the successor function, i.e.,
$s(x)$ is the successor of $x.$ Let 
$$E_n=\{p(0), p(s^2(0)),...,p(s^{2n}(0))\}.$$
 There are two inductive learning algorithms to produce
two different theories for $E_n.$

{\bf Case 1.} One inductive learning algorithm produces $T_n=\{p(0); p(s^2(x))\leftarrow p(x)\},$ given $E_n.$ Then $T_n$ is a Horn logic program and the least Herbrand model of $T_n$ is 
$$M_n=\{p(0),p(s^2(0)),...,p(s^{2m}(0)),...\}.$$
Then we have that 
$$\begin{array}{l}
T=\lim_{n\rightarrow \infty} T_n=T_1;\\
M=\lim_{n\rightarrow \infty} M_n=M_1.
\end{array}$$
The least Herbrand model of $T$ is $M.$

{\bf Case 2.} Another inductive learning algorithm produces $S_n=\{p(s^{2n}(0)); p(x)\leftarrow p(s^2(x))\},$ given $E_n.$
 Then, $S_n$ is a Horn logic program and the least Herbrand model, say
$N_n$ of $S_n$ is $E_n.$ But
$$\begin{array}{l}
S=\lim_{n\rightarrow\infty} S_n=\{p(x)\leftarrow p(s^2(x))\};\\
N=\lim_{n\rightarrow\infty} N_n=\lim_{n\rightarrow\infty} E_n=M_1.
\end{array}$$
The least Herbrand model, say $M(S),$ of $S$ is equal to
the empty set. Then
$$M(S)\ne N.$$
That is, for any $e\in N,\ S\not\vdash e.$

\bigskip

{\bf Example 3.1.} Assume that $E_n$ is given as above, then the GOLEM system produces the following sequence of the Horn logic programs:
$$\begin{array}{rl}
T_0&=\{p(0)\};\\
T_1&=\{p(0); \pi_1\};\\
T_2&=\{p(0); \pi_1; \lgg(\pi_1, \pi_2)\}\\
&=\{p(0); \pi_1, p(s^2(x))\leftarrow p(x)\}\\
&=\{p(0); p(s^2(x))\leftarrow p(x)\},\\
T_3&=T_2,\\
&\cdots\\
T_n&=T_2,\\
&\cdots
\end{array}$$
where
$$\begin{array}{rl}
\pi_2&=\{ \lnot(\lnot p(0)\lor p(s^2(0))), p(s^4(0))\}\\
&=\{p(0), p(s^4(0))\}\land \{\lnot p(s^2(0)), p(s^4(0))\},\\
\pi_1&=\{\lnot p(0),p(s^2(0))\},\\
\lgg(\pi_1,\pi_2)&=\{p(s^2(x)), \lnot p(x)\}.
\end{array}$$

We change the ordering of the occurrences of $p(s^{2n}(s))$'s and see what the GOLEM system gets.

{\bf Example 3.2.} Assume that 
$$\begin{array}{rl}
E'_0&=\{p(s^4(0))\},\\
E'_1&=E_0\cup \{p(s^2(0))\},\\
E'_2&=E_1\cup \{p(0)\},\\
& \cdots 
\end{array}$$
$$\begin{array}{rl}
E'_{3k}&=E_{3k-1}\cup \{p(s^{6k+4}(0))\},\\
E'_{3k+1}&=E_{3k}\cup \{p(s^{6k+2}(0))\},\\
E'_{3k+2}&=E_{3k+1}\cup \{p(s^{6k}(0))\},\\
&\cdots
\end{array}$$
Then the GOLEM system produces the following sequence of the Horn logic programs:
$$\begin{array}{rl}
S_0&=\{p(s^4(0))\};\\
S_1&=\{p(s^4(0)); \gamma_1\};\\
S_2&=\{p(s^4(0));\gamma_1; \lgg(\gamma_1,\gamma_2)\}\\
&=\{p(s^4(0)); \gamma_1; p(x)\leftarrow p(s^2(x))\}\\
&=\{p(s^4(0)); p(x)\leftarrow p(s^2(x))\},\\
&\cdots\\
S_{3k+i}&=\{p(s^{6k+4-2i}(0)); p(x)\leftarrow p(s^2(x))\},\\
&\cdots
\end{array}$$
where
$$\begin{array}{rl}
\gamma_1 &=\{\lnot p(s^4(0)), p(s^2(0))\},\\
\gamma_2 &=\{p(s^4(0)), p(0)\}\land \{\lnot p(s^2(0)), p(0)\},\\
\lgg(\gamma_1,\gamma_2)&=\{p(x), \lnot p(s^2(x))\}.
\end{array}$$

By the above discussion, this example shows that the GOLEM system is not limit-correct, and sensitive to the ordering of the examples.

\bigskip

\noindent{\bf 4. The prioritized GOLEM system}
\smallskip

In the section, we give the prioritized GOLEM system which is limit-correct and not sensitive to the ordering
of the examples. In [6] the authors proposed 
an assumption on the Horn logic programs
which guarantees that the least Herbrand model of the limit of a sequence of Horn logic programs is the limit of the
least Herbrand models of the logic programs, and the least Herbrand model of the former limit is stable with respect to
the sequences of the Horn logic programs. 

{\bf Definition 4.1.} A clause is {\it simple} if every subterm occurring
in the body of the clause occurs in the head of the clause. A logic program is {\it simple} if every clause in it is 
simple. 

The assumption in section 1 requires that the Horn logic program be simple.

Given an example set $E$, we modify the GOLEM system to be a new alogrithm $\calP$ such that $\calP(E)$ is simple.
\bigskip

We first define a relation $\prec$ on the literals. Given two literals $l$ and $l',$ we say that $l$ has a higher
priority than $l',$ denoted by $l\prec l',$ if every sub-term occurring in $l$ occurs in $l'.$ 

{\bf Proposition 4.2.} $\prec$ is a pre-order, that is, $\prec$ is reflexive and transitive.

\bigskip

{\bf The prioritized GOLEM system:}
\begin{quote}{\tt
Suppose we are given a logic program $\Gamma$ (i.e., a background knowledge) and two examples (two ground atoms) $E_1$ 
and $E_2$ such that $\Gamma\not\vdash E_1$ and $\Gamma\not\vdash E_2.$ We construct the lgg $C$ of $E_1$ and $E_2$ relative 
to $\Gamma,$ so $\Gamma\land C\vdash E_1\land E_2$ and $C$ is used only once in the derivation of both $E_1$ and $E_2.$ 
Let $\Gamma=\{p_1,...,p_n,...\}.$ 

Define $C_1=((\lnot q_1\lor \lnot q_2\lor \cdots)\lor e_1),  C_2=((\lnot q'_1\lor \lnot q'_2\lor \cdots)\lor e_2),$ 
set $C=\lgg(C_1,C_2),$ where $q_1, q_2, ..., e_1$ is the set $\Gamma\cup \{E_1\}$ under the order $\prec,$ that is,
$\{q_1,q_2,...,e_1\}=\Gamma\cup \{E_1\}$ and $e_1\not\prec q_i$ for every $i;$ similarly define $\{q'_1, q'_2,...,e_2\}.$ }
\end{quote}

\bigskip

{\bf The prioritized GOLEM system on sequences:}

\begin{quote}{\tt
Given a sequence $\{E_n\}$ of example sets, at stage $n,$ input $E_n.$ If $E_n\not\prec E_i$ for any $i<n$ then use
the GOLEM system directly to produce a Horn logic program, say $\calP(E_n);$ otherwise, find the least $i<n$ such that $E_n\prec E_{i+1},$ then
use the GOLEM sytem to $\bigcup_{j>i}^n E_j$ with background knowledge $\calP(E_i).$ Let $\calP(E_n)$ be the resulted theory.}
\end{quote}

\bigskip

{\bf Theorem 4.3.} Given an example set $E,\ \calP(E)$ is simple.

{\sl Proof.} By the definition of the pre-order, when a clause $\pi$ is enumerated in $\calP(E)$ the instance of the head of $\pi$ always
has the lowest priority among $E.$ This guarantees that $\pi$ is simple.
\bigskip

Therefore, we have the following theorem.

{\bf Theorem 4.4.} $\calP$ is convergent, and limit-correct. That is, given an increasing sequence $\{E_n\}$ of positive example sets, 

(4.1) $\{\calP(E_n)\}$ is a sequence of Horn logic programs which has the set-theoretic limit;

(4.2) $\lim_{n\rightarrow\infty} \calP(E_n)$ is limit-correct with respect to 
$\lim_{n\rightarrow\infty}E_n.$ 

{\sl Proof.} By the discussion in section 1, we only need to prove (4.1). To prove (4.1), by the definition of the pre-order
$\prec,\ \prec$ is well-founded. Given an example $e,$ there are only finitely many $e'$ with $e'\prec e.$ Assume that
when $e$ is enumerated
in $E_n$ for some $n,$ a clause $\pi$ is produced. $\pi$ is extracted out of $\calP(E_n)$ only when an example $e'\prec e$ is 
enumerated in $E_{n'}$ for some $n'> n.$ Hence, $\pi$ cannot be enumerated in $\calP(E_n)$  and be extracted out of 
$\calP(E_n)$ for infinitely many $n.$ Therefore, $\{\calP(E_n)\}$ has the set-theoretic limit.

\bigskip

\noindent{\bf 5. Conclusion and further work}

\smallskip
When input a set $E$ of examples, the prioritized GOLEM system produces a Horn logic program $\calP(E)$ which is simple.
The simple Horn logic programs have many good properties that a common Horn logic program has not. The prioritized GOLEM
system is based on the syntactical properties of the examples, that is, the priority order $\prec$ on the literals, which
make the prioritized GOLEM system useful in diverse applications.

A further work could be based on the distance defined on the terms or formulas of a logical language. Such the distance
definitions can be the ones given by Fitting([3]) or Nienhuys-Cheng([9]). Then the Cauchy sequences of terms or formulas can
be defined. Then the convergence and the limit-correctness can be defined in terms of the Cauchy sequences of
the example sets and the Horn logic programs. It is conjectured that the prioritized GOLEM system also satisfies 
the convergence and the limit-correctness defined on Cauchy sequences. 

\bigskip
\bigskip

\noindent{\bf References:}

\begin{description}

\item{[1]} F.~Bergadano and D.~Gunetti, {\it Inductive Logic Programming: from machine learning to software engineering,} The MIT Press, Cambridge, Massachusetts, London, 1996.

\item {[2]} M.~Dahr, {\it Deductive Databases: Theory and Applications},
International Thomson Computer Press, 1997.

\item{[3]} M.~Fitting, Metric methods, three examples and a theorem, {\it J. of Logic Programming} 21(1994), 113-127.

\item {[4]} W.~Li, An Open Logic System, {\it Science in China} (Scientia
Sinica) (series A), 10(1992)(in Chinese), 1103-1113.

\item {[5]} W.~Li, A logical Framework for Inductive Inference and Its
rationality, in Fu,N.(eds.): {\it Advanced Topics in Artificial Intelligence,%
} LNAI 1747, Springer, 1999.

\item {[6]} J.~W.~Lloyd, {\it Foundations of Logic Programming},
Springer-Verlag, Berlin, 1987.

\item{[7]} S.~Ma, Y.~Sui and K.~Xu, The limits of the Horn logic programs. {\it Proc. of 18th 
International Conference on Logic Programming} (poster session), Denmark, to appear, 2002. Full
paper is available at http://www.nlsde.buaa.edu.cn/\~{ }kexu

\item{[8]} S.~Muggleton and C.~Feng, Efficient inductive of logic programs, in: 
{\it Proc. of the First Conf. on Algorithmic Learning Theory,} Tokyo, 1990, Ohmsha.

\item {[9]} S.~H.~Nienhuys-Cheng, Distances and limits on Herbrand
Interpretations, {\it Proc. of the 8th International Workshop on Inductive
Programming}, LNAI 1446, Springer, 1998, 250-260.

\item{[10]} S.~H.~Nienhuys-Cheng, Distance between Herbrand Interpretations: 
A Measure for Approximations to a Target Concept, in: {\it Proc. of the 7th International
Workshop on Inductive Logic Programming}, LNAI 1297, Springer, 1997,
213-226. 

\item{[11]} G.~Plotkin, A note on inductive generalization, in B.~Meltzer and D.~Michies, eds., {\it Machine Intelligence} 5, 153-163, Edinburgh Univ. Press, 1970.

\item{[12]} G.~Plotkin, A further note on inductive generalization, in B.~Meltzer and D.~Michies, eds., 
{\it Machine Intelligence} 6, 101-124, Edinburgh Univ. Press, 1971.

\end{description}
\end{document}